\documentclass{article}
\usepackage{spconf,amsmath,graphicx,epstopdf,cite}
\usepackage{algorithm}
\usepackage{algorithmic}
\DeclareMathOperator\argmin{arg\,min}

\graphicspath{{Fig/}}

\title{A SCALABLE CONVOLUTIONAL NEURAL NETWORK
FOR TASK-SPECIFIED SCENARIOS VIA KNOWLEDGE DISTILLATION}
\name{Mengnan Shi$^{\dagger}$ \qquad Fei Qin$^{\dagger \star}$ \qquad Qixiang Ye$^{\dagger}$ \qquad Zhenjun Han$^{\dagger}$ \qquad Jianbin Jiao$^{\dagger}$\thanks{This research is supported in partial by Nature Science Foundation of China under Grant No.61401426, 61271433, 61671427, Beijing Municipal Science and Technology Commission under Grant No.Z161100001616005, and Science and Technology Innovation Foundation of Chinese Academy of Sciences under Grant No.CXJJ-16Q218.}}
\address{$^{\dagger}$School of Electronics, Electrical and Communication Engineering \\
University of Chinese Academy of Sciences, Beijing, China.\\$^{\star}$fqin1982@ucas.ac.cn}

\begin{document}
%
\maketitle
\begin{abstract}
In this paper, we explore the redundancy in convolutional neural network, which scales with the complexity of vision tasks. Considering that many front-end visual systems are interested in only a limited range of visual targets, the removing of task-specified network redundancy can promote a wide range of potential applications. We propose a task-specified knowledge distillation algorithm to derive a simplified model with pre-set computation cost and minimized accuracy loss, which suits the resource constraint front-end systems well. Experiments on the MNIST and CIFAR10 datasets demonstrate the feasibility of the proposed approach as well as the existence of task-specified redundancy. 
\end{abstract}
\section{Introduction}
\label{sec:intro}

Visual algorithms in front-end systems play an important role to boost related industry, e.g., augmented reality systems in daily lives, vehicle recognition systems in transportation systems, and unmanned aerial vehicle (UAV) surveillance systems for security purpose. It is generally expected that the implementation of such systems will bring modern society a more convenient life.

This brilliant future has been guaranteed by the recent popular deep learning framework in computer vision community, represented by convolutional neural networks (CNN), with their significant performance, which also play as the chevaux-de-frise role for the front-end implementation. This is due to the fact that most deep algorithms are only optimized for the accuracy performance. As a result, these algorithms require thousands of GPU cores to process an image in real-time, and even more GPU cores to train a model in days or weeks. Although training cost can be ignored for front-end systems, the computation cost for the test stage is still too high to be afforded by most embedded systems without powerful GPUs. This is why the currently famous alpha-Go was really a huge chassis deployed in UK rather than in the game site.

While some researchers are working with the ‘deeper and deeper’ fashion to keep improving the accuracy of deep learning framework \cite{he2016deep,simonyan2014very1,szegedy2015going}, some others have noticed the redundancy of deep framework, i.e., most network components negligibly contribute to the overall performance. Several state-of-the-art works \cite{anwar2015fixed1,han2015learning,hinton2015distilling,luo2016face,polyak2015channel,romero2015fitnets,sun2016sparsifying,lebedev2015fast,han2015deep,ba2014deep1,anwar2015structured,kalinowski2015compact} have shown significant potential performance to derive a simplified CNN without accuracy loss in various usages. The works of \cite{han2015learning,polyak2015channel,sun2016sparsifying,hassibi1993second,lecun1989optimal} follow the pruning style, which propose different metrics to evaluate the contribution of network components. According to these contribution metrics, low response components, named the shared redundancy, will be disabled. Some other researchers \cite{hinton2015distilling,luo2016face,romero2015fitnets,papamakarios2015distilling} propose to use knowledge distillation, which employs the acquired knowledge of the original model to train another, usually smaller, model. The smaller model, named student model, will mimic the original model, named teacher model, with comparable performance but higher efficiency.

Beyond these works, most front-end deployed visual algorithms are task-specified. For example, an UAV based surveillance system only focuses on the military targets and won't be interested in a cat or cup. But most popular convolutional models are learned from big data with thousands of object classes, the knowledge among which is more than necessary for these scenarios. Existing work \cite{agrawal2014analyzing} has demonstrated that the filters in a network have different responses to different targets. It is straightforward to make the hypothesis that the redundancy of CNN scales with the complexity of a given task, e.g., if the categories of interested targets significantly decreases, there will be corresponding redundancy, named task-specified redundancy, in the network, which can be further removed. 

We propose a task-specified knowledge distillation algorithm to derive a simplified model. The knowledge distillation method relies on transferring the learned discriminative information from a teacher model to a student model. We first analyze and demonstrate the redundancy of the neural network related to \emph{a priori} complexity of the given task. We then train a student model by re-defining the loss function from a subset of the relaxed target knowledge according to the task information. The new model can satisfy the constraints of both the computation cost and residue accuracy. 

The organization of this paper is as follows: section 2 provides the problem description, proposed methodology is discussed in section 3, and section4 describes the experiment design and results analyses. The conclusion of this work as well as potential future work is discussed in section 5.

\section{Problem description}
\label{sec:format}

A typical CNN in vision applications takes an image as input and processes it into a feature vector for image classification, scene classification, object detection, and object tracking, etc. It is worth noting that the core aim of CNN is to learn features, the capability of which is usually hard to be quantitatively determined. A plausible method is to employ the accuracy of classification task as an instead metric. This is reasonable, since the usage of classification is embedded in most visual algorithms as an essential tool. The accuracy of classification can be treated as a rough metric to express the capability of feature representation \cite{agrawal2014analyzing,krizhevsky2012imagenet}.

As shown in fig.1, the input of a convolutional layer is a set of feature maps. The size of input of a convolutional layer is $C_{i}*I_{s}$, where $C_{i}$ is the number of input channels, and $I_{s}$ specifies the size of input feature maps. The output of the convolutional layer is $C_{o}*O_{s}$ feature maps, where $C_{o}$ is the number of output channels and $O_{s}$ the size of output feature maps. The process is implemented by convolutional operation with $C_{o}*C_{i}$ filters of size $K_{s}$. If the size of input and other convolutional parameters have been set, the computation complexity of a convolutional layer can be expressed as:
\begin{equation}
\vspace{-0.2cm}
\begin{array}{l}
	CP_{conv} \sim O(\sum_{conv_{layer}}(C_{i}*C_{o}*K_{s}))
\end{array}
\vspace{0.2cm}
\end{equation}

The situation of the full connected layers will be slightly different, but still scaled with $N_{i}$ and $N_{o}$, i.e., the number of input neurons and output neurons:
\begin{equation}
\vspace{-0.2cm}
\begin{array}{l}
	CP_{fc} \sim O(\sum_{fc_{layer}}(N_{i}*N_{o}))			
\end{array}
\vspace{0.2cm}
\end{equation}
\begin{figure}[htb]
  \centering
  \centerline{\includegraphics[width=8cm]{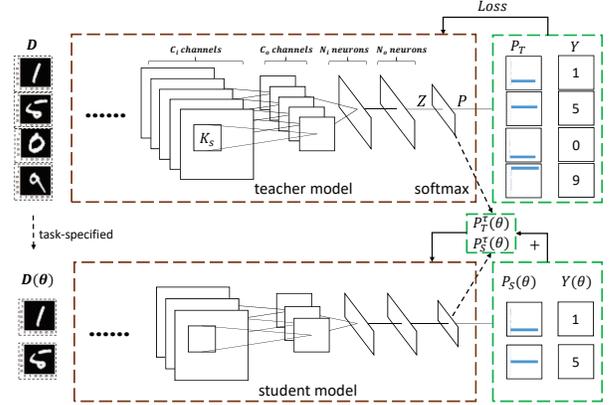}}
  \caption{The scheme of task-specified knowledge distillation}\medskip
\label{Fig:5}
\end{figure}

This architecture leads to extremely high computation cost, say in billions of instructions in processor. Processing a standard size image in the popular AlexNet requires 1.5 billion floating point operations \cite{han2015learning}. As reported \cite{han2015learning,polyak2015channel,sun2016sparsifying}, network components in a CNN model like connections, filters, and channels, may have different contributions to the overall feature representation. Remaining those with high contribution and pruning low contribution ones will only bring tiny influences to the overall accuracy. It is assured that the redundancy in CNN models can be reduced by avoiding unimportant connections, filters and channels \cite{han2015learning,polyak2015channel,sun2016sparsifying}, as:
\begin{equation}
\vspace{-0.2cm}
\begin{array}{l}
	\mathop{\underset{C_{i},C_{o},N_{i},N_{o}}\argmin}(CP_{conv}+CP_{fc}),\\
\ \ subject \ to: \ acc_{loss} >= acc_{orig}-acc
\end{array}
\vspace{0.2cm}
\end{equation}
where $acc$ represents the accuracy of compressed model, and $acc_{orig}$ represents the accuracy of original model. $acc_{loss}$ is the threshold. This can be understood to minimize the model complexity, and to guarantee the negligible accuracy loss in the compressed model. 

Many popular CNNs with extraordinary performance have been trained with large datasets, like ImageNet with millions of samples covering thousands of classes. Nevertheless, in practical front-end systems, only few classes of targets will be interested in. By removing the redundancy representing these uninterested targets, the efficiency of new network could be further increased. Under this hypothesis, the problem should be extended to reduce task complexity related redundancy:
\begin{equation}
\vspace{-0.2cm}
\begin{array}{l}
	\mathop{\underset{C_{i},C_{o},N_{i},N_{o}}\argmin}(CP_{conv}+CP_{fc}),\\
\ \ subject \ to: \ Tacc_{loss} >= Tacc_{orig}-Tacc
\end{array}
\vspace{0.2cm}
\end{equation}
where $Tacc$ represents the accuracy of compressed model on specified task, and $Tacc_{orig}$ represents the accuracy of original model on specified task. $Tacc_{loss}$ is the threshold.

\section{Methodology}
\label{sec:pagestyle}

Most recently, Hinton \emph{et al.} \cite{hinton2015distilling} pointed out that the original cumbersome teacher models can be utilized to improve the performance of small student models by a transfer learning method, knowledge distillation.

In knowledge distillation, an additional, if not included in the original model, softmax layer is required at the top of CNN, which converts the original feature vectors, i.e. logits $Z$, into probabilities for each class:
\begin{equation}
\vspace{-0.2cm}
\begin{array}{l}
	p_{i}^{\tau}=\frac{exp⁡(z_{i}/\tau)}{\sum_{j}exp⁡(z_{j}/\tau)}
\end{array}
\vspace{0.2cm}
\end{equation}
where $p_{i}^{\tau}$ is the $i$th element of probability distribution vector $P^{\tau}$; $z_{i}$ is the $i$th element of logits vector $Z$; $j$ sums all the classes; $\tau$ is a relaxation parameter, i.e. setting $\tau$ equal to 1 will be back compatible with normal CNN model. With the increase of $\tau$, the probability distributions will be softened. The softened probability vector $P^{\tau}$ contains the relative similarity of the input class to other classes, which reveals how the CNN models discriminate and generalize between given classes. Although the softmax layer is must in knowledge distillation, the application does not have to use the softmax vector, e.g., they can still feed feature vectors of full-connected layer to the original classifier.

Let the original model be defined as the teacher model $T$ and the expected small one as the student model $S$. Let $P_{S}$ be the output probabilities of the student model. Let $Y$ be the true label vector of a sample. $P_{T}^{\tau}$ from the teacher model is named as the soft targets for the student model. Let $P_{S}^{\tau}$ be the softened probabilities from the student model with the same relaxation $\tau$.
The cross entropy function is shown as follow:
\begin{equation}
\vspace{-0.2cm}
\begin{array}{l}
	H(P,Q)=-\sum_{i}p_{i}log(q_{i})
\end{array}
\vspace{0.2cm}
\end{equation}
where $p_{i}$ is the $i$th element of vector $P$, and $q_{i}$ likewise. 

Aided by the soft targets from teacher model, the loss function in knowledge distillation is defined as:
\begin{equation}
\vspace{-0.2cm}
\begin{array}{l}
	L_{KD} = \frac{1}{N}\sum_{N}((1-\lambda)H(Y,P_{S})+\lambda H(P_{T}^{\tau},P_{S}^{\tau}))
\end{array}
\vspace{0.2cm}
\end{equation}
where $N$ is the number of samples and $\lambda$ is a tunable parameter to balance both cross entropies. By the introducing of $\lambda H(P_{T}^{\tau},P_{S}^{\tau})$, knowledge distillation helps the student model to learn how the original model treats the sample. Existing works \cite{hinton2015distilling,romero2015fitnets} have shown that compared with direct training, the student model converges with higher accuracy similar to the original one.

The extension of knowledge distillation based solution into the task scalable version is more than convenient. Set the dataset to train teacher model as $D$ and the dataset to train student model as $D(\theta)$ named as the transfer set, which has less task complexity than former, and typically is a sub set of $D$, containing only task of interested targets. As shown in fig.1, a subset of corresponding soft targets can be obtained as $P_{T}^{\tau}(\theta)$. This has been shown in algorithm 1 as a variation form, since in the real implementation, this action is equal to directly acquiring the $P_{T}^{\tau}(\theta)$ from $D(\theta)$, which aims to avoid the intermedia step of $P_{T}^{\tau}$. Similarly, we could obtain the task-specified version of $Y(\theta)$.

Now, the task-specified student model $S(\theta)$ will be trained with an modified loss function $L_{KD}(\theta)$:
\begin{equation}
\vspace{-0.2cm}
\begin{array}{l}
L_{KD}(\theta) = \frac{1}{N}\sum_{N}((1-\lambda)H(Y(\theta),P_{S}(\theta))\\
\ \ \ \ \ \ \ \ \ \ \ \ \ \ \ \ \ \ \ \ \ \ \ \ \ \ \ \ \ \ \ + \lambda H(P_{T}^{\tau}(\theta),P_{S}^{\tau}(\theta)))
\end{array}
\vspace{0.2cm}
\end{equation}			    
where $P_{S}(\theta)$, and $P_{S}^{\tau}(\theta)$ are the outputs of student model in each iteration. The notation $\theta$ is to show the fact that they will ideally only response to $D(\theta)$.

The detailed algorithm has been provided in the following table:
\begin{algorithm}[h]
\renewcommand{\algorithmicrequire}{\textbf{Input:}}
\renewcommand\algorithmicensure {\textbf{Output:} }
\caption{\label{Algorithm_1} Task-specified Knowledge Distillation}
\begin{algorithmic}[1]
\REQUIRE $\boldsymbol{D}$, $\boldsymbol{T}$, $\boldsymbol{\theta}$
\ENSURE $\boldsymbol{S(\theta)}$
\STATE add a softmax layer to $\boldsymbol{T}$
\STATE use $\boldsymbol{D}$ to train the teacher model $\boldsymbol{T}$
\STATE use $\boldsymbol{T}$ to capture soft targets $\boldsymbol{P_{T}^{\tau}(\theta)}$ from each sample in $\boldsymbol{D(\theta)}$
\STATE set the architecture of the student model $\boldsymbol{S(\theta)}$
\STATE train the student model with soft targets $\boldsymbol{P_{T}^{\tau}(\theta)}$ and $\boldsymbol{D(\theta)}$, iteratively until the accuracy converges \\
\begin{center}
$\boldsymbol{S(\theta)=\mathop{\underset{S(\theta)}\argmin}L_{KD}(\theta)}$
\end{center}
\end{algorithmic}
\end{algorithm}

Noting that, as the original model T may already exist, or even be equipped with a softmax layer as classifier, this algorithm may have some variations starting from different steps. It should also be noted that, as the size, or more precisely the architecture, of student model was manually deigned. It could be in the various forms, e.g. we could even remove a layer or add a layer if we would like to do so. As discussed in \cite{he2016deep,simonyan2014very1,szegedy2015going}, the architecture not the size of parameters will decide the performance of the deep framework, for example a thinner but deeper network may bring better performance. The knowledge distillation method will secure the possibility to utilize the potential architecture gain in the future.
\section{EXPERIMENTS}
\label{sec:typestyle}

We conducted our experiments on two classification datasets, MNIST and CIFAR10. All the codes were deployed in a desktop PC with one K80 GPU card. The architecture of the teacher model for MNIST is 20-50-500-10, 
\begin{figure*}[htb]
\centering
\begin{minipage}[b]{0.32\linewidth}
  \centering
  \centerline{\includegraphics[width=5cm]{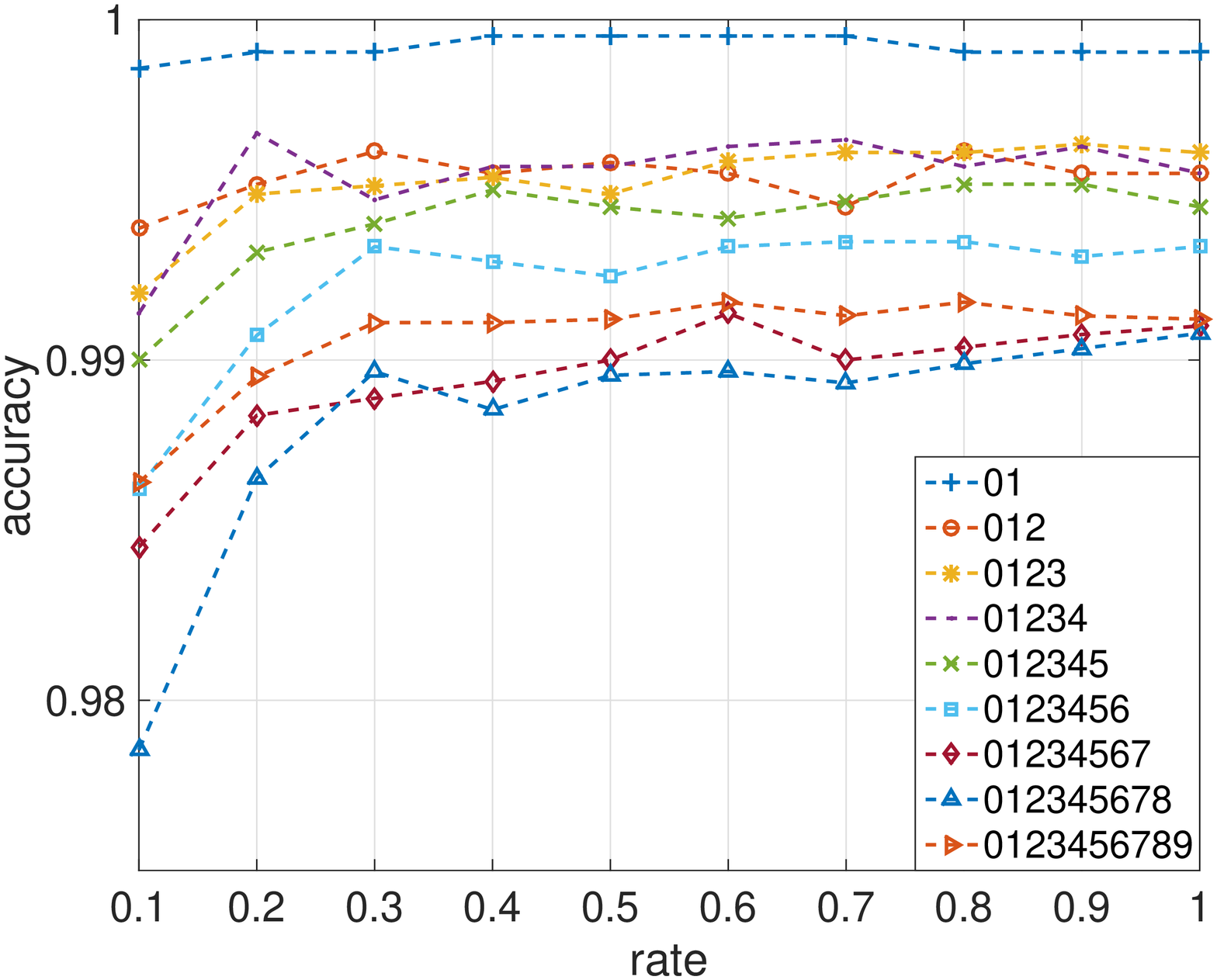}}
  \centerline{(a)}\medskip
\end{minipage}
\begin{minipage}[b]{0.32\linewidth}
  \centering
  \centerline{\includegraphics[width=5cm]{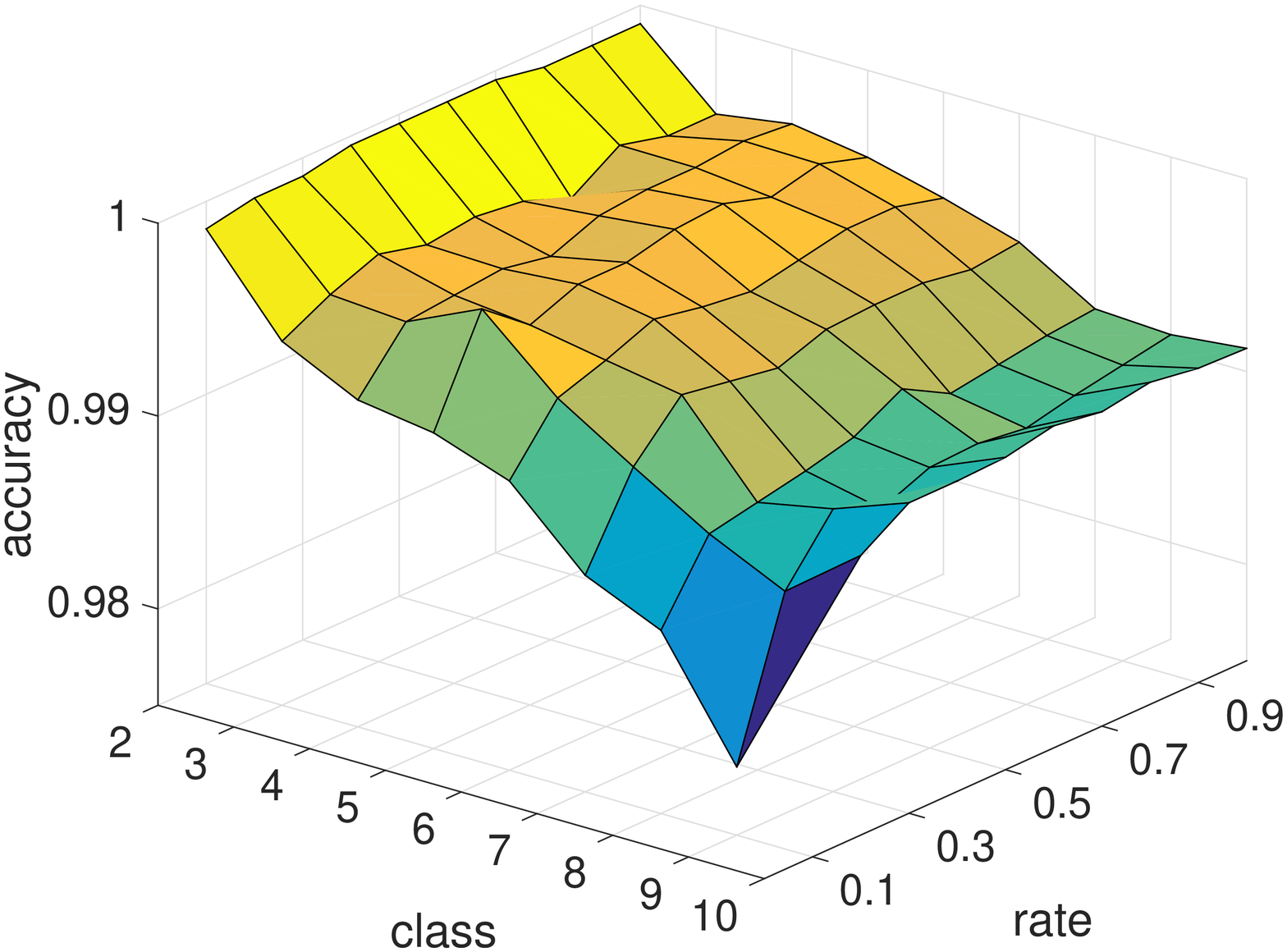}}
  \centerline{(b)}\medskip
\end{minipage}
\begin{minipage}[b]{0.32\linewidth}
  \centering
  \centerline{\includegraphics[width=5cm]{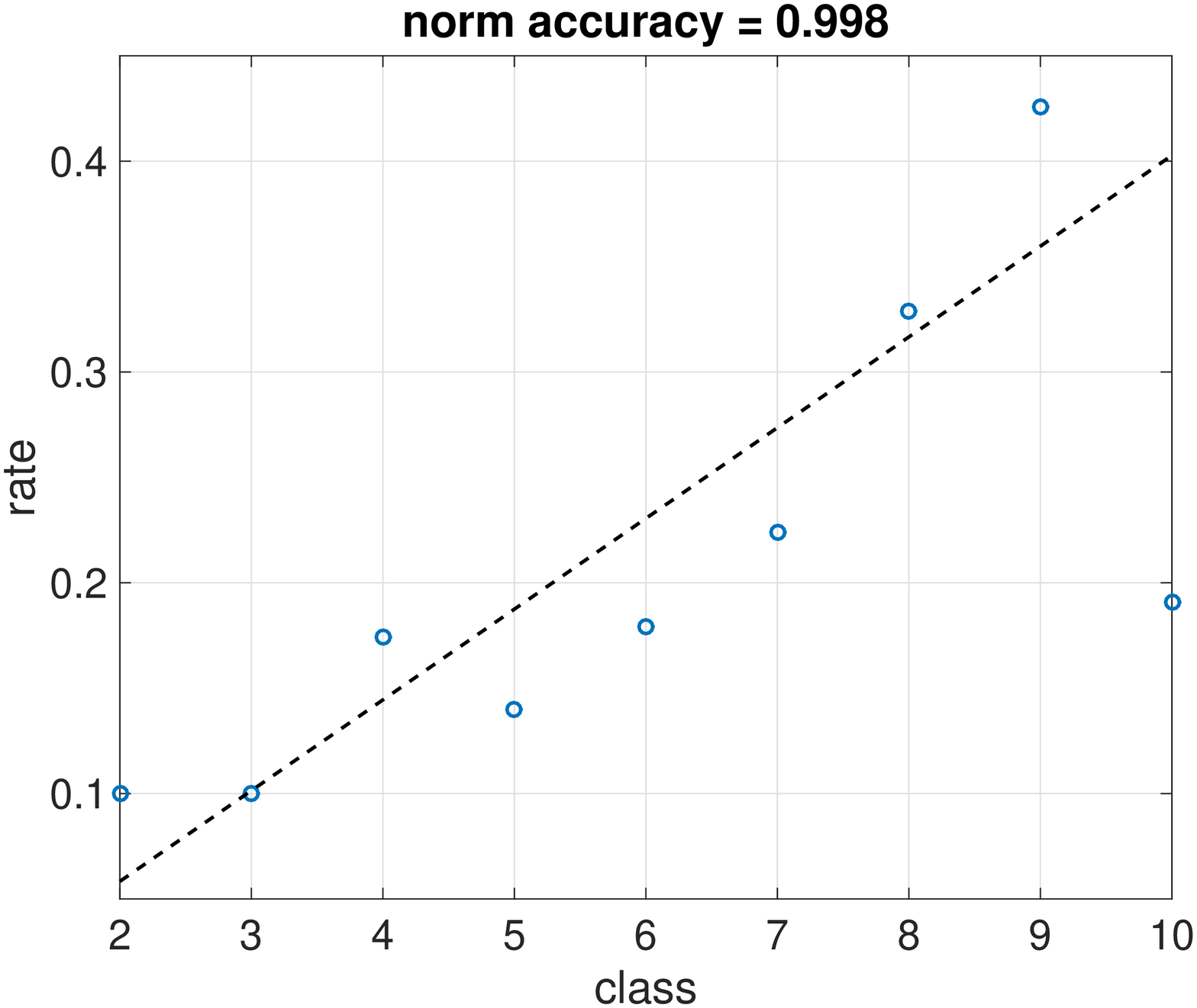}}
  \centerline{(c)}\medskip
\end{minipage}
\vspace{-0.3cm}
\caption{(a) Results of MNIST; (b) re-drawed 3D version; (c) relationship between rate and class given normalized accuracy.}
\vspace{-0.3cm}
\label{Fig:3}
\end{figure*}
\begin{figure*}[htb]
\centering
\begin{minipage}[b]{0.32\linewidth}
  \centering
  \centerline{\includegraphics[width=5cm]{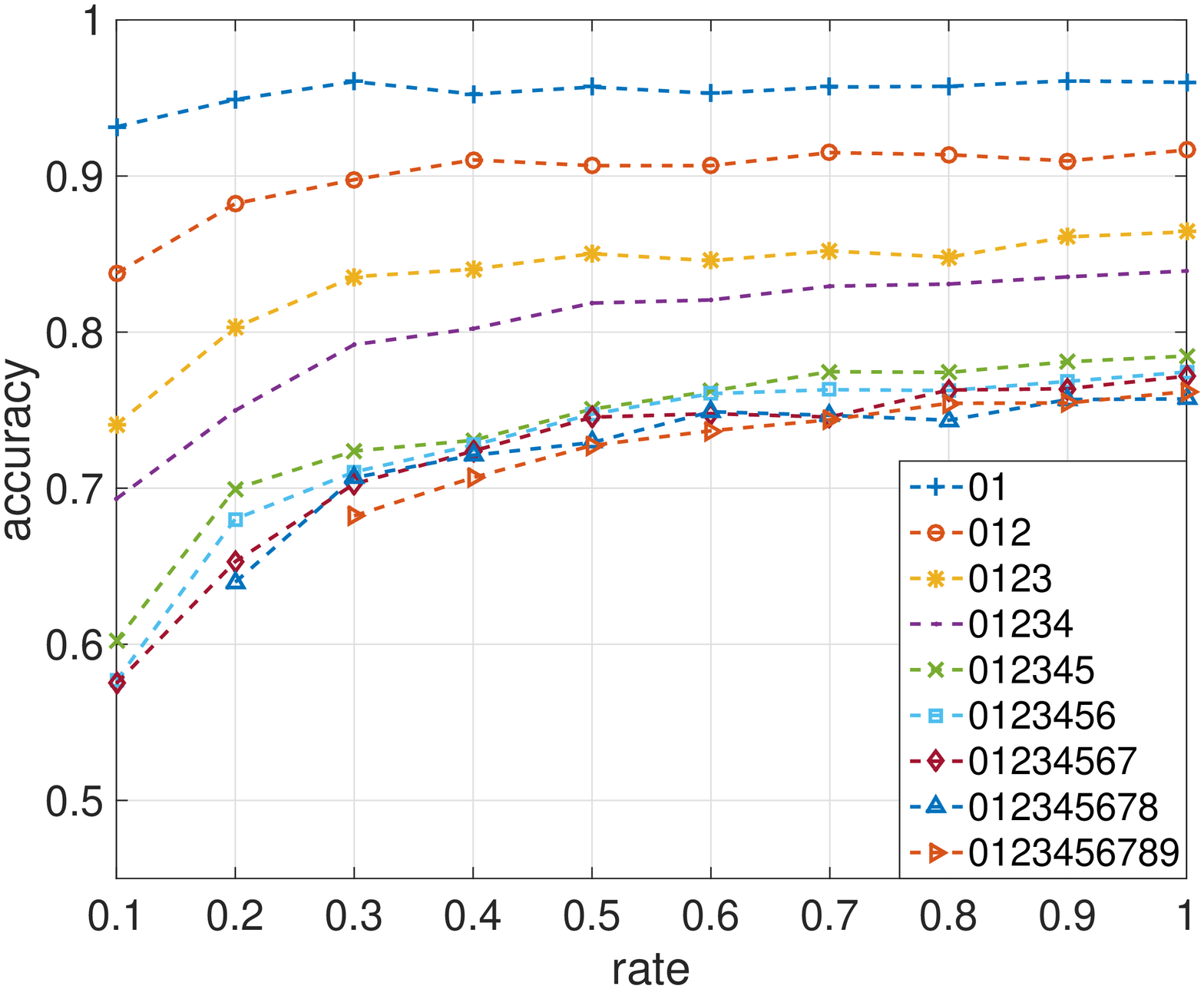}}
  \centerline{(a)}\medskip
\end{minipage}
\begin{minipage}[b]{0.32\linewidth}
  \centering
  \centerline{\includegraphics[width=5cm]{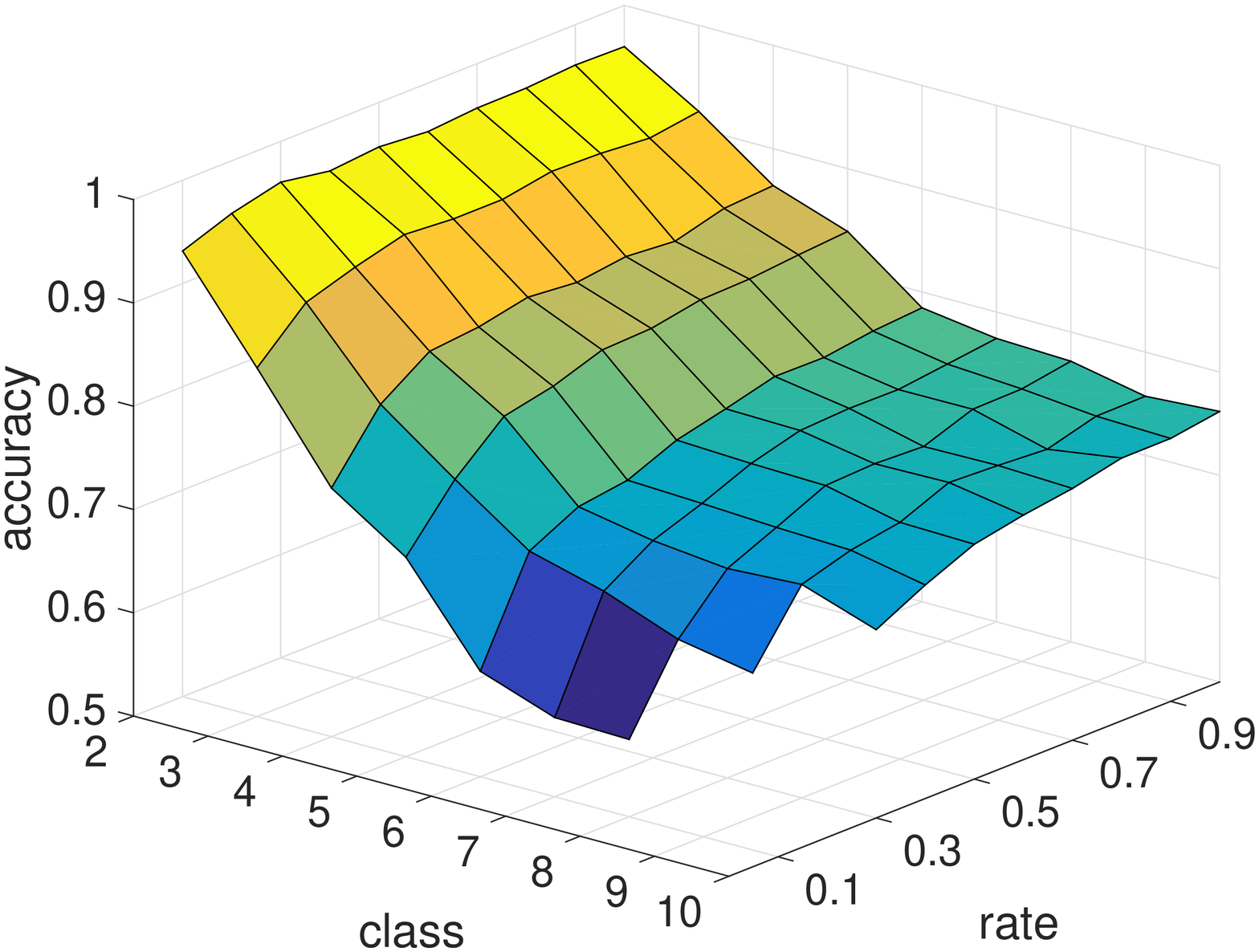}}
  \centerline{(b)}\medskip
\end{minipage}
\begin{minipage}[b]{0.32\linewidth}
  \centering
  \centerline{\includegraphics[width=5cm]{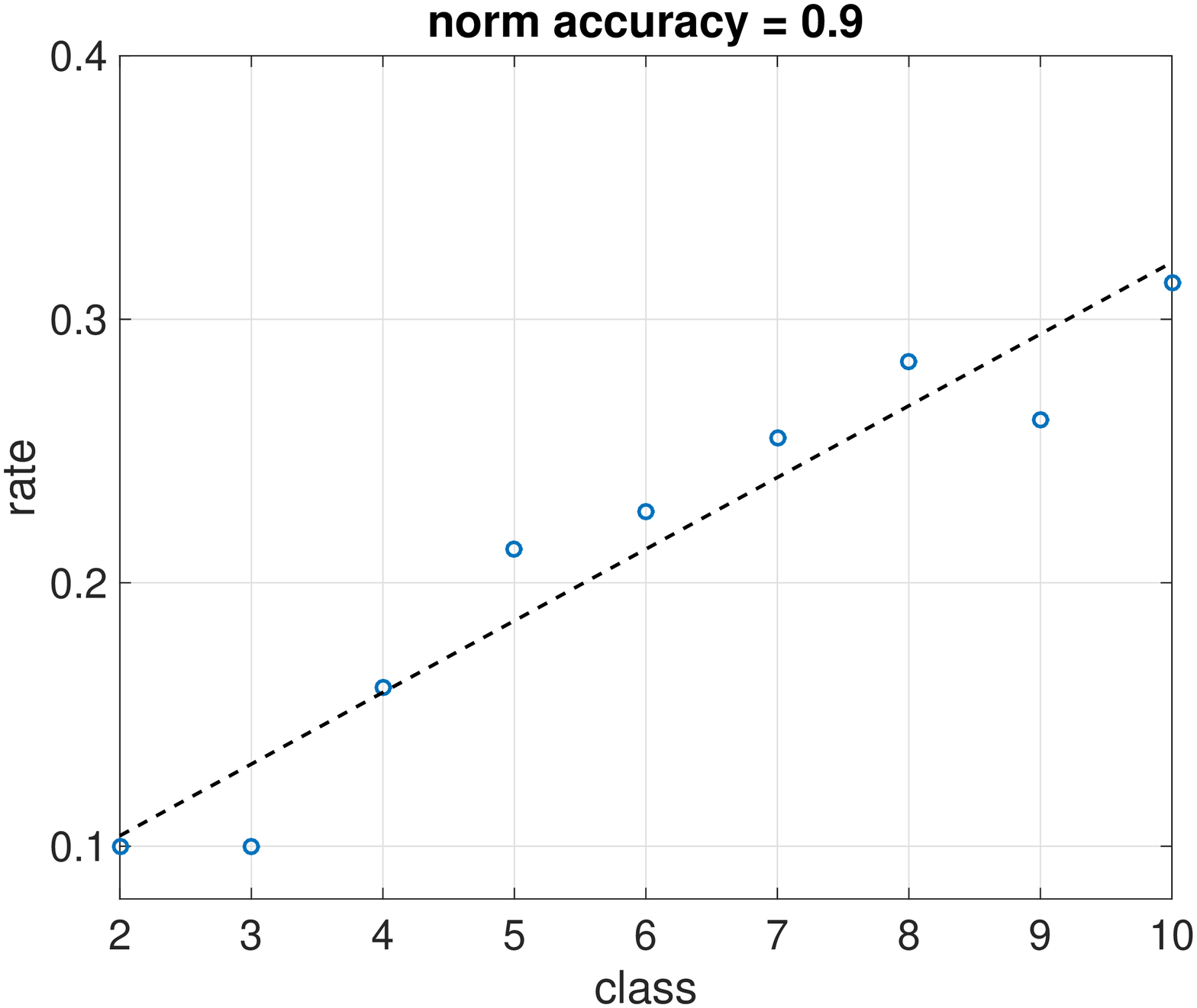}}
  \centerline{(c)}\medskip
\end{minipage}
\vspace{-0.3cm}
\caption{(a) Results of CIFAR10; (b) re-drawed 3D version; (c) relationship between rate and class given normalized accuracy.}
\vspace{-0.3cm}
\label{Fig:3}
\end{figure*}
where the first two layers are convolutional and last two are full-connected. Similarly, the architecture used for CIFAR10 is 32-32-64-10, where the first three layers are convolutional and the last one is full-connected. For both models, outputs of last full-connected layer are treated as inputs of the softmax layer. In the experiments $\tau$=3 is selected, although it works similarly in a larger range within [3:10]. $\lambda$ is carefully selected as 0.9 to observe system performance with emphasis on transfer effect, i.e. transfer learning contributes more than true label training.

In the experiments, 10 classes of datasets will be divided into subsets to demonstrate the performance of task-specified student model. We set our transfer sets according to the following strategy. Take MNIST for example. The training dataset $D$ for the teacher model covers the whole training samples, say all ten classes. And we set transfer datasets covering from two classes to nine classes, from $D(\theta_{2})$ to $D(\theta_{9})$. Let $D(\theta_{2})$ cover samples with label `0' and `1', $D(\theta_{3})$ cover `0', `1' and `2', which will guarantee that $D(\theta_{2})\subseteq D(\theta_{3})\cdots\subseteq D$. Under these assumptions, the difficulty of sub tasks will be monotonically increasing.

The architecture of student model is manually designed in knowledge distillation. In the experiments, all layers except the last full-connected layer and the softmax layer are compressed with the same rate. For example, if the compression rate for MNIST model is 0.5, the structure is 10-25-250-10. All the retrained student models will be tested to show the performance of proposed algorithm.

Fig.2(a) and fig.3(a) are the direct results of MNIST and CIFAR10. The results brief the basic fact that the accuracy decreases with the size of model. Moreover, those compressed models for simpler task suffer less accuracy loss. As fig.2(a) shown, in the results of MNIST, the accuracy of the simplest task $D(\theta_{2})$ drops 0.05 percent while the accuracy for the original $D$ drops 0.48 percent, both with size scale of 0.1. Similarly, the accuracy for $D(\theta_{2})$ drops 2.85 percent while the accuracy for $D(\theta_{8})$ drops 19.69 percent for CIFAR10 with the same size scale of 0.1 (``rate'' in the figure). These results illustrate that for lower task complexity, the models have more redundancy, which can be further compressed. Fig.2(b) and fig.3(b) are the re-drawed 3-D version of the experiment results to show a global performance change in the experiments.
 
From fig.2(b), and fig.3(b), we could derive fig.2(c) and fig.3(c), which provide an intuitive validation to our hypothesis. The performance of different $D(\theta)$ has been normalized, which shows the relative performance drop. Then, the estimated model sizes to achieve the same performance drop, i.e. a threshold, have been provided in fig.2(c) and fig.3(c). Easy to notice that for the threshold of 99.8\%, a simpler task $D(\theta_{2})$ could remove more redundancy than the original D for MNIST, while the situation for CIFAR10 is likewise.

\section{CONCLUSION \& FUTURE WORK}
\label{sec:majhead}

In this paper, we first propose the hypothesis that the redundancy among CNN models or other similar deep learning systems will correspond to task complexity, which means potential performance gain especially for the front-end deployed vision systems. We design the algorithm from the recently proposed knowledge distillation to further reduce task-specified redundancy. Experiments on MNIST and CIFAR10 have been provided to validate our hypothesis and algorithm. Due to the page limitation, detailed experiments will be provided in our future work.

Obviously, knowledge distillation benefits from its architecture of an \emph{a priori} model size, which is critical to resource constraint embedded systems. But at the same time, the re-trained student model may still have shared redundancy, i.e., the low contribution network components. A following pruning stage may be employed to further improve the performance in our future work. Also as discussed in section 4, the task complexity may not be monotonically scaled with the interested targets number. A carefully designed metric to better describe the task complexity will be examined as well.

\bibliographystyle{IEEEbib}
\bibliography{strings,refs}

\end{document}